\documentclass[sigconf, nonacm]{acmart}

\usepackage{amsmath,amssymb,amsfonts}
\usepackage{algorithmic}
\usepackage{graphicx}
\usepackage{textcomp}
\usepackage{xcolor}

\usepackage{hyperref}
\usepackage{xspace}
\usepackage{subcaption}
\usepackage{booktabs}
\usepackage{float}
\usepackage{multirow}
\usepackage{multicol}
\usepackage{array}
\usepackage{tabularx}
\usepackage{xfrac}

\usepackage{enumitem}

\usepackage{comment}

\usepackage{circledsteps}
\newcommand\myCircled[2][]{\ifmmode%
\Circled[fill color=black,inner color=white,#1]{\footnotesize\mathsf{#2}}%
\else%
\Circled[fill color=black,inner color=white,#1]{\footnotesize\sffamily#2}%
\fi%
}

\begin{document}

\title{\textsc{FusionSense}: Tri-Stage Near-Sensor Learning \\for Runtime-Adaptive Multimodal Edge Intelligence}
\renewcommand{\shorttitle}{\textsc{FusionSense}: Tri-Stage Near-Sensor Learning for Runtime-Adaptive Multimodal Edge Intelligence}



\author{Sanggeon Yun}
\affiliation{%
  \institution{University of California, Irvine}
  \city{Irvine, CA}
  \country{USA}}
\email{sanggeoy@uci.edu}

\author{Ryozo Masukawa}
\affiliation{%
  \institution{University of California, Irvine}
  \city{Irvine, CA}
  \country{USA}}
\email{rmasukaw@uci.edu}

\author{Minhyoung Na}
\affiliation{%
  \institution{Kookmin University}
  \city{Seoul}
  \country{South Korea}}
\email{minhyoung0724@kookmin.ac.kr}

\author{Hyunwoo Oh}
\affiliation{%
  \institution{University of California, Irvine}
  \city{Irvine, CA}
  \country{USA}}
\email{hyunwooo@uci.edu}

\author{Yoshiki Yamaguchi}
\affiliation{%
  \institution{Shibaura Institute of Technology}
  \city{Saitama}
  \country{Japan}}
\email{bp21016@shibaura-it.ac.jp}

\author{Wenjun Huang}
\affiliation{%
  \institution{University of California, Irvine}
  \city{Irvine, CA}
  \country{USA}}
\email{wenjunh3@uci.edu}

\author{SungHeon Jeong}
\affiliation{%
  \institution{University of California, Irvine}
  \city{Irvine, CA}
  \country{USA}}
\email{sungheoj@uci.edu}

\author{Mohsen Imani}
\affiliation{%
  \institution{University of California, Irvine}
  \city{Irvine, CA}
  \country{USA}}
\email{m.imani@uci.edu}







\begin{abstract}
Autonomous systems and smart-industry deployments increasingly split computation across near-sensor, edge, and cloud resources, where tight energy, latency, and reliability budgets demand \emph{run-time} adaptivity. In practice, deciding \emph{what} to compute and transmit at each point is pivotal; yet as multimodal sensor suites (cameras, LiDAR/depth, etc.) proliferate at the edge, most prior approaches either (i) fuse modalities on powerful servers or (ii) apply uni-modal near-sensor filters that ignore cross-modal dependencies, leading to redundant transmissions or missed events.
We present \textsc{FusionSense}, a fusion-aware intelligent sensing framework for energy-constrained autonomous edge systems. Lightweight near-sensor classifiers are trained via a three-step procedure: (i) a server-side fusion model learns the downstream task, (ii) \emph{filter-out-safe} (FoS) labels quantify each modality's necessity relative to the fused decision, and (iii) an edge-side fusion model is compacted by injecting near-sensor predictions as auxiliary signals. The result is a run-time decision layer that jointly reduces compute and communication while scaling linearly with sensor count.
On a dual-modality (RGB{+}Depth/LiDAR) setup with SynDrone, \textsc{FusionSense} sustains task quality at substantially higher data-reduction rates than uni-modal filters and delivers large end-to-end gains: up to 33$\times$ lower energy at 1\% FoI prevalence, 11$\times$ at 10\%, a 92.3\% reduction in quality loss at a fixed 30\% data reduction, and roughly 1.5$\times$ higher energy savings than the best prior filtering baseline.

\end{abstract}

\maketitle

\section{Introduction}\label{sec:introduction}
Autonomous systems in modern manufacturing increasingly split workloads across near-sensor/on-device processing, edge nodes, and cloud back-ends, where platforms must make run-time decisions under tight energy, latency, reliability, and thermal constraints~\cite{alikhani2024seal,isuwa2023content,taufique2025hidp,yun2025missiongnn,yun2025loghd}. As multimodal sensors (e.g., RGB, LiDAR/depth, IMU) become standard at the edge, it is essential to model inter-sensor correlations \emph{before} transmission and fusion; otherwise, systems waste uplink bandwidth and risk accuracy drops when complementary cues are required~\cite{alikhani2023dynafuse}. While large back-end models often exploit such cross-modal structure, edge devices must also internalize it to remain both efficient and accurate~\cite{taufique2025hidp,xun2024fluid}. In practice, \emph{intelligent sensing}—pre-processing at or near the sensor—has been adopted in agriculture~\cite{liakos2018machine}, healthcare~\cite{bahri2018big}, logistics~\cite{woschank2020review}, privacy~\cite{pramanik2023overview}, and security~\cite{khalid2023improved}; however, most deployments are uni-modal and task-specific, and thus fail to capture the cross-modal dependencies.

As the number of edge devices increases, the amount of data that must be processed at servers also increases, leading to severe scalability issues. A promising direction is near-sensor adaptation~\cite{yun2024hypersense, huang2024plug}, where tiny AI models positioned at sensors filter redundant data so that only task-relevant information is relayed for server-side operations. This strategy is effective when the probability of observing a \emph{frame of interest} (FoI) is low, conserving energy with modest quality loss. \emph{Yet,} when modern edge nodes integrate \emph{multiple} sensors, simply applying independent uni-modal filters becomes suboptimal: it (i) transmits redundant modalities even when a subset suffices for the fused decision, and (ii) may over-filter and miss events that require complementary cues. Moreover, data volume and compute grow with each added modality, shifting scalability and thermal challenges to the edge itself.

We propose \textsc{FusionSense}, a fusion-aware intelligent sensing framework that follows a pragmatic design principle. We treat multi-sensor streams as correlated random variables and train lightweight near-sensor classifiers to approximate the joint decision boundaries of a more complex server-side fusion model. 
Our methodology is informed by multimodal fusion research~\cite{sharma2022recent, kim20223d} and validated empirically. In particular, \textsc{FusionSense} introduces a \emph{three-step} learning process that (1) trains a high-capacity server fusion model for the downstream task, (2) derives \emph{filter-out-safe} (FoS) labels that indicate when each modality is unnecessary for the fused decision, and (3) compacts an edge-side fusion model by injecting near-sensor predictions as auxiliary signals—yielding a run-time decision layer that co-optimizes compute and communication across the edge–cloud continuum and scales \emph{linearly} with the number of sensors.

We instantiate and evaluate \textsc{FusionSense} on a dual-modality camera–LiDAR setup. Experiments demonstrate that \textsc{FusionSense} maintains task quality at substantially higher data-reduction rates than uni-modal filters and unlocks large end-to-end gains: up to 33$\times$ lower energy at 1\% FoI prevalence and 11$\times$ at 10\%; at a fixed 30\% data reduction, we observe a 92.3\% reduction in quality loss relative to a uni-modal filter; and compared with the best prior filtering baseline, our method yields roughly 1.5$\times$ higher energy savings.

Our novel work provides the following contributions:
\vspace{-1mm}\begin{itemize}[leftmargin=*]
    \item We present \textsc{FusionSense}, a \emph{fusion-aware intelligent sensing} framework that reduces total energy usage of multimodal fusion systems by learning near-sensor keep/drop decisions conditioned on the fused task—addressing a growing gap in multimodal edge sensing for smart-industry autonomy.
    \item \textsc{FusionSense} introduces a \textit{three-step training} method that (i) learns a server fusion model, (ii) derives modality-wise FoS supervision from the fused decision, and (iii) compacts an edge fusion model via near-sensor predictions, enabling deployment on power-constrained devices across the compute continuum.
    \item On camera–LiDAR experiments, \textsc{FusionSense} demonstrates up to 33$\times$ lower energy at 1\% FoI and 11$\times$ at 10\%, a 92.3\% reduction in quality loss at 30\% data reduction, and roughly 1.5$\times$ higher energy savings than the best prior filtering baseline—while scaling \emph{linearly} with sensor count.
\end{itemize}

\vspace{-2mm}
\section{Related Work}\label{sec:background}
\subsection{Multi-modal Sensor Fusion}

Understanding complex environments often requires fusing heterogeneous sensor streams so that complementary cues can be jointly exploited for perception and decision making~\cite{sharma2022recent}. Recent progress in multimodal representation learning further shows that diverse modalities can be embedded into shared spaces to improve recognition robustness and transfer~\cite{girdhar2023imagebind}. In autonomous perception, camera–LiDAR fusion has become a cornerstone for reliable 3D understanding under diverse operating conditions~\cite{sato2023revisiting}. Concretely, feature-level and bird’s-eye-view pipelines such as BEVFusion~\cite{liu2023bevfusion}, 3D Dual-Fusion~\cite{kim20223d}, EA-LSS~\cite{hu2023ea}, and UFO~\cite{kim2024ufo} report strong results on large-scale benchmarks like nuScenes~\cite{Caesar_2020_CVPR}. These systems typically presume that all modalities (or suitably preprocessed features) are available to a central fusion model; their primary objective is to maximize downstream accuracy and robustness given full-modality inputs. In contrast, our perspective emphasizes the \emph{pre-fusion} stage—deciding \emph{what} to transmit or compute before fusion—so that multimodal benefits can be retained while reducing end-to-end resource usage.

\vspace{-2mm}
\subsection{Energy Efficient Edge/AIoT Computing}\label{sec:efficientedgecomputing}

Deep neural networks have delivered state-of-the-art results in vision, audio, and beyond, and are increasingly deployed on mobile/embedded platforms~\cite{9061142,7740096,yun2025hyperdimensional,yun2025decohd}. This trend has catalyzed numerous AIoT applications spanning smart environments and monitoring~\cite{li2022emotional,teng2021aiot,madhusudhan2024blockchain,yun2026contextual}. To fit tight energy and latency envelopes, classic efficiency levers include quantization, pruning, and specialized low-power hardware~\cite{sun2023review}. A complementary line explores \emph{near-sensor intelligence}, where lightweight models suppress or summarize raw data so only task-relevant information is forwarded downstream~\cite{yun2024hypersense,huang2024plug}. However, in distributed settings the communication subsystem (e.g., Wi-Fi/5G) can dominate energy and latency~\cite{7488250,yang2023breaking}, meaning that sending “everything, but compressed”~\cite{8848858,9473014} may still be suboptimal when informative events are sparse or when certain modalities are temporarily unhelpful. These observations motivate \emph{fusion-aware} decisions that co-optimize compute and communication by conditioning transmission on the contribution of each modality to the fused task, rather than applying modality-agnostic compression or uni-modal filtering.

\vspace{-2mm}
\subsection{Intelligent Sensing over Dynamic Multimodal Data}

Real-world operation exposes sensors to modality-specific degradations (e.g., low light, glare, fog) that can impair individual streams and shift which modalities are most informative~\cite{linnhoff2022measuring}; in parallel, security concerns such as LiDAR spoofing highlight the need for redundancy and graceful degradation~\cite{sato2023revisiting}. A rich literature therefore studies dynamic fusion and robustness: learning when and how to trust each modality~\cite{liu2017learning,tsai2018learning,takahashi2019deep,zhi2020factorized}. Notably, the Crossmodal Compensation Model (CCM) detects corrupted inputs and compensates in a self-supervised manner~\cite{9561847}, while DynMM uses mixture-of-experts routing across modality combinations to cut computation without sacrificing accuracy~\cite{Xue_2023_CVPR}. Yet, most of these methods assume that all modalities (or features) have already traversed the uplink to a central model; they optimize \emph{post-acquisition} fusion rather than \emph{pre-fusion} acquisition/transmission policies. Our work is complementary: we leverage the fused model’s decision to supervise lightweight near-sensor selectors that act \emph{before} transmission, thereby avoiding unnecessary data movement while preserving the benefits of dynamic, robust multimodal fusion downstream.

\begin{figure*}[t!]
  \centering
  \includegraphics[width=\linewidth]{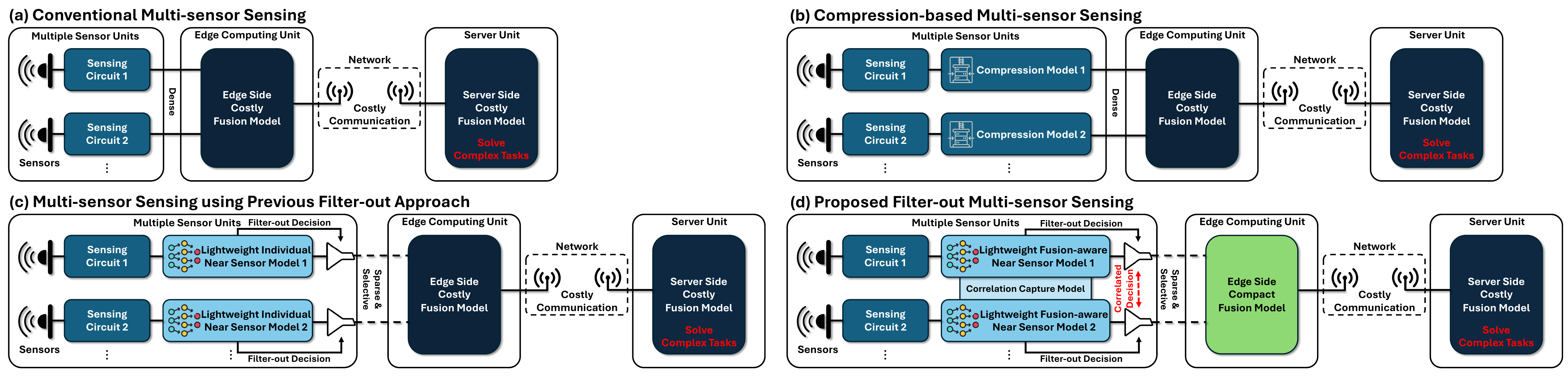}
  \vspace{-9mm}
  \caption{\small Comparison of our proposed sensing and information processing pipeline with other approaches: (a) Conventional approach, (b) Compression-based approach, (c) Using a previously proposed filter-out approach designed for a single sensor environment, and (d) ours.}
  \label{Fig:Ours_diagram}
    \vspace{-5mm}
\end{figure*}

\vspace{-2mm}
\section{Intelligent Multi-Sensing Design}\label{sec:intelligent_multisensing_design}

Complex machine learning tasks often require substantial models that are challenging to deploy on edge devices. For example, a cutting-edge transformer-based classification model~\cite{9746312} required 80 hours of training on four NVIDIA Tesla V100 GPUs, even though it was designed to use fewer computational resources than its predecessors. Many studies also involve hefty models for intricate tasks, such as large pre-trained multi-modality models~\cite{wu2022wav2clip} and substantial transformer models~\cite{baade2022mae}. As a result, implementing these advanced deep learning-based multi-modal fusion tasks in real time on edge sensors poses significant practical difficulties. Our approach addresses these challenges by binarizing the tasks, specifically by detecting only the essential \textit{"frame of interest"} (FoI) data from multiple sensors needed for complex operations. Unlike earlier methods that used near-sensor modules, our strategy achieves optimal energy savings through a design tailored for a fusion environment. In this section, we present our intelligent multi-sensing framework, compare it with prior methods in a fusion context, and explain our model training technique.

\vspace{-2mm}
\subsection{Intelligent Multi-Sensing Framework}

Our proposed approach is essentially different from previous approaches~\cite{8848858, huang2024plug, yun2024hypersense} since it is specifically designed for multi-sensor fusion scenarios. Comparison with previous approaches: (a) Conventional approach without near-sensor paradigm, (b) Compression-based approach where it compresses sensed data before transmitting them, (c) Using a previously proposed filter-out approach designed for single sensor environment, and (d) ours, using diagrams is shown in \autoref{Fig:Ours_diagram}. In the case of the conventional approach, it sends out all the sensed data from multiple sensors, leading to excessive data transmission and processing resulting in extreme energy inefficiency when the probability of encountering a frame of interest is low. Next, the compression-based approach uses a near-sensor paradigm by placing compression modules near each sensor. Although it can reduce communication costs between different modules by the compression rate showing better energy efficiency compared to the conventional approach, cost from computation is still exhibited and merely contributes to the energy efficiency in the low FoI probability scenario. Another filter-out-based approach specifically designed for the low FoI probability cases, resolves the issue by placing near-sensor models for each sensor. Although this approach achieved promising energy efficiency in a single-sensor scenario, in a multi-sensor scenario where we need multiple near-sensor models for each sensor, it is inappropriate to naively implement the approach to achieve the maximum possible energy efficiency with low-quality loss resulting from miss filter-out due to its individuality unaware of relationships between different modalities in the fusion model.

To keep the system deployable on real-world edge devices, we do not require specialized hardware beyond standard accelerators (e.g., Google Edge TPU or NVIDIA Jetson) for near-sensor models. Scalability is addressed by the linear addition of near-sensor modules per sensor and a single centralized fusion process at either the edge or server. Thus, from a hardware perspective, the design neither multiplies complexity nor forces sophisticated networking conditions.

\vspace{-2mm}
\subsection{Three-step Model Training}

\begin{figure*}[t!]
  \centering
  \includegraphics[width=0.9\linewidth]{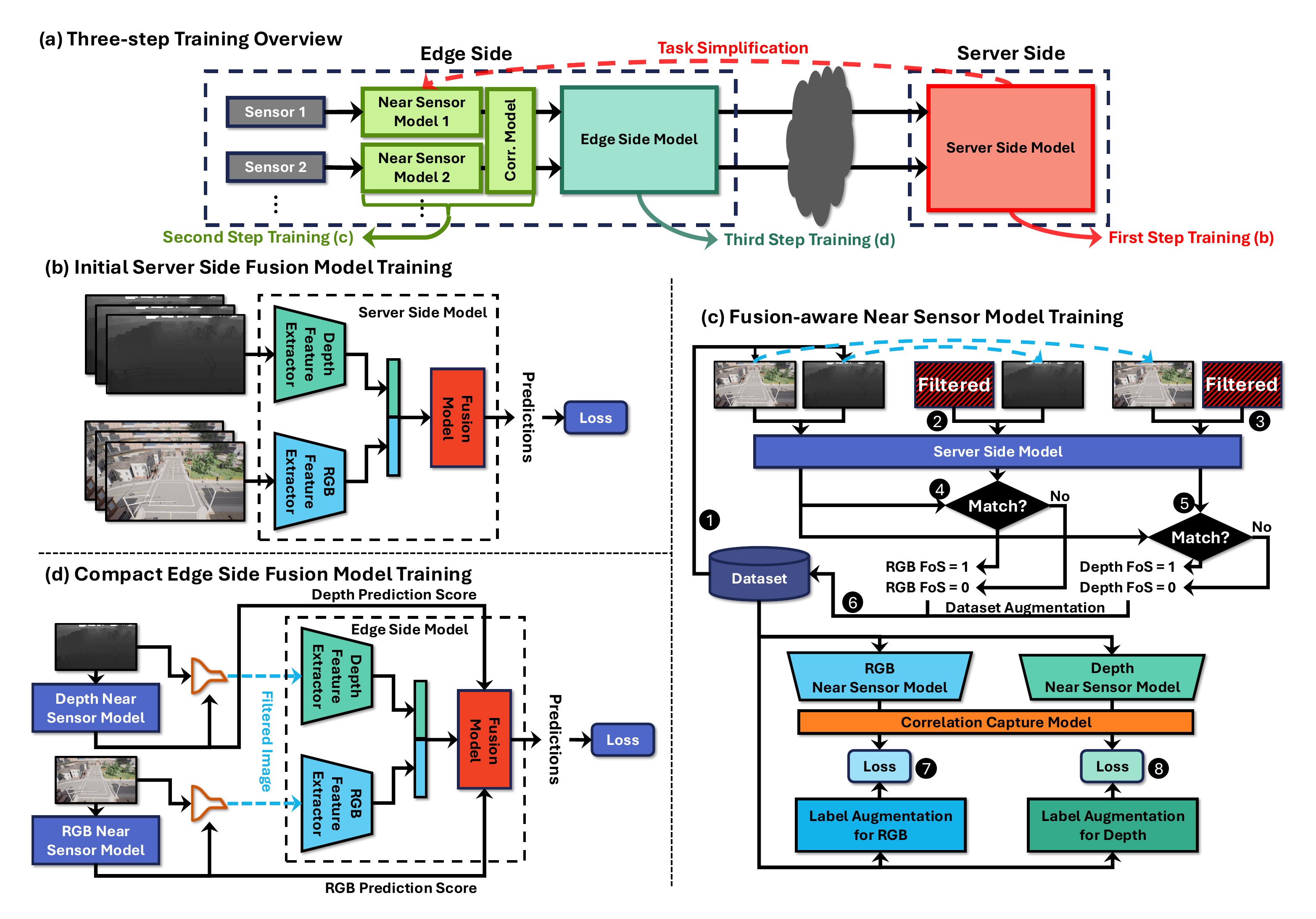}
  \vspace{-6mm}
  \caption{\small Overview of the Proposed \textit{Three-step Training} Method: (a) presents a schematic representation of the entire Three-step Training process. The process begins with the initial training phase depicted in (b), proceeds to the secondary training phase illustrated in (c), and concludes with the tertiary training phase outlined in (d).}
  \label{Fig:Threestep_training_diagram}
    \vspace{-5mm}
\end{figure*}

\begin{table}[]
\caption{\small Truth table or decision table of label augmentation for near-sensor models training. Labels for RGB and Depth near-sensor models (right side) are decided by the three binary values: FoI, RGB FoS, and Depth FoS (left side).}
\label{tab:FoS_truthtable}
\vspace{-4mm}
\centering
\resizebox{0.7\columnwidth}{!}{
\begin{tabular}{ccc|cc}
\toprule
        FoI & RGB FoS & Depth FoS & RGB Label & Depth Label \\ \midrule
        0 & 0 & 0 & 0 & 0 \\
        0 & 0 & 1 & 0 & 0 \\
        0 & 1 & 0 & 0 & 0 \\
        0 & 1 & 1 & 0 & 0 \\
        1 & 0 & 0 & 1 & 1 \\
        1 & 0 & 1 & 0 & 1 \\
        1 & 1 & 0 & 1 & 0 \\
        1 & 1 & 1 & 0 & 1 \\
\bottomrule
\end{tabular}%
}
    \vspace{-6.5mm}
\end{table}

\autoref{Fig:Threestep_training_diagram} illustrates our proposed three-step training across from near sensor, edge side to server side. The overall view of three-step training in terms of edge-server architecture view is shown in \autoref{Fig:Threestep_training_diagram}.(a). In this architecture, there are three types of machine learning models to train: i) near-sensor models, ii) an edge-side fusion model, and iii) a server-side fusion model. We first start with training the server-side fusion model which solves complex tasks requiring heavy machine learning models as shown in \autoref{Fig:Threestep_training_diagram}.(b). Here, we use a late fusion model that fuses different modalities by combining corresponding feature embeddings from each separated feature extraction model. In our implementation, each sensor (RGB or depth) has a dedicated backbone CNN, and a final fully connected fusion layer merges these features. The training is standard supervised learning with multi-label classification. However, it is not necessary to follow the late fusion design since there are no constraints on the server-side model, where its model architecture can be varied by the task to conduct, in our proposed three-step training. We target to have near-sensor models each aware of correlations between other modalities in this server-side fusion model. This is achieved by conducting the next training step which is illustrated in \autoref{Fig:Threestep_training_diagram}.(c). 

To train fusion-aware near-sensor models, it takes a data augmentation approach by introducing \textit{Filter-out Safe} or FoS of each modality. If FoS is $1$ for a certain modality, it indicates even if we do not give that modality data as input to the server-side fusion model or filter that modality, the output or decision made from the server-side model will not be changed. Otherwise, where FoS is 0, the output or decision made from the server-side fusion model will be changed. We anticipate that this FoS will effectively capture cases where certain modalities may either provide duplicated information already contained in other modalities or contain irrelevant data due to severe weather conditions affecting certain sensors etc. This capability allows the system to disregard such modalities, thereby reducing redundancy. In order to have FoS for each modality, we first retrieve the decision from the server-side fusion model when we give all of the modalities (\myCircled{1}). Next, we also retrieve the decision of the server-side fusion model when we filter out RGB data (\myCircled{2}) and also when we filter out depth data (\myCircled{3}). The filter-out is done by giving zeroed data with the same dimensionality as the original input data of the corresponding modality. Then we compare the output from the original input data with the outputs from filtered-out input data. FoS for RGB input data is set to be 1 if the output from the original input and the output from \myCircled{2} is the same and if not, it is set to be 0 (\myCircled{4}). For the depth modality, the same procedure is applied except we use depth filtered-out case instead of RGB filtered-out case to decide FoS for depth (\myCircled{5}). These FoSs for RGB and depth modality are retrieved for all data points, augmenting the dataset (\myCircled{6}). Finally, near-sensor models for RGB (\myCircled{7}) and depth modality (\myCircled{8}) are trained by label augmentation using the newly augmented FoS information. The label augmentation for each modality is conducted using three pieces of information: FoI, RGB FoS, and Depth FoS, as shown in \autoref{tab:FoS_truthtable}.

Lastly, compactization of the server-side fusion model to deploy on the edge side is conducted which is illustrated in \autoref{Fig:Threestep_training_diagram}.(d). It is important to make the edge-side fusion model as compact as possible considering the practical purpose where edge devices are required to operate without outside energy sources. It is achieved by introducing prediction scores of each near-sensor model to the edge-side late fusion model. We assumed that introducing prediction scores from the near-sensor models to the late fusion model can weakly and one directionally -- from near-sensor models to the fusion model -- combine the fusion model with the near-sensor models resulting in the fusion model behaving like a larger model. Based on this assumption, we train a fusion model following the same architecture as the server-side fusion model but smaller size using filtered modality data with the prediction scores from the near-sensor models.

\vspace{-3mm}
\section{Experiments}\label{sec:experiments}

\begin{figure*}[t!]
    \centering    \includegraphics[width=0.65\linewidth]{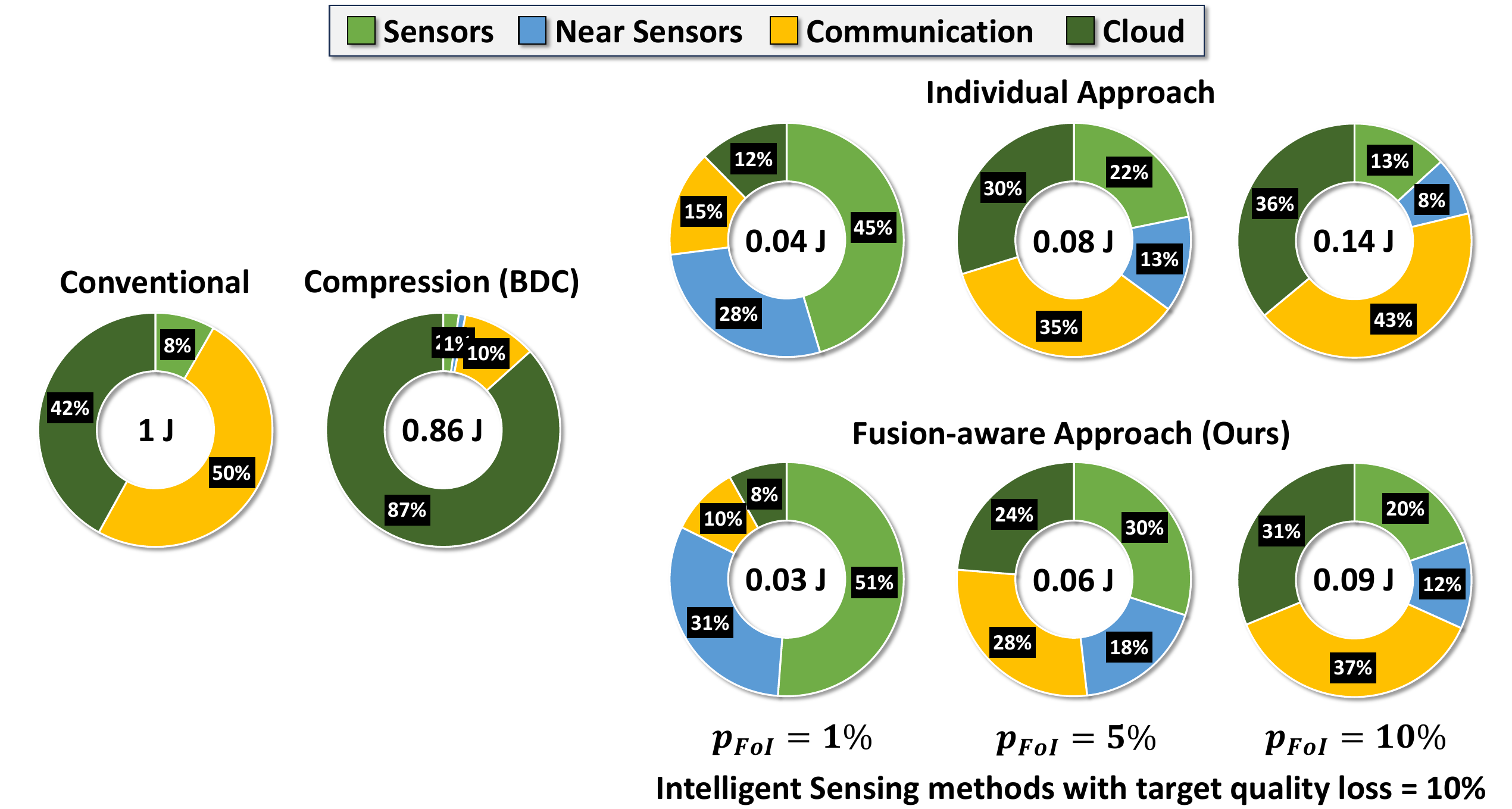}
    \vspace{-4mm}
    \caption{\small Comparative distribution of energy consumption across four methods: the conventional method, the compressive near-sensor approach, the previous filtering-out approach using individual near-sensor models, and our proposed method, across varying probabilities of FoI. The total energy consumption values are normalized to the total of the conventional method and displayed at the center of each distribution.}
    \label{fig:pie_breakdown}
    \vspace{-5mm}
\end{figure*}

\vspace{-0.8mm}
\subsection{Experimental Setup}
We implemented and executed our framework with both a software framework and GPU accelerators. Specifically, the implementation was completed by using PyTorch and NumPy which support CNN layers and classification. In model training, the Adam optimizer, exponential learning rate scheduler with $\gamma = 0.95$, and binary cross entropy loss were used. We used 60 epochs in total and selected the best model based on a 10\% validation split from the training set, and we performed a grid search on small sets of hyperparameters (batch sizes of 8, 16, 32 and learning rates of $10^{-3}$ or $10^{-4}$). For feature extractors, we used MobileNetV3~\cite{koonce2021mobilenetv3} for near-sensor models and RegNet 400mf~\cite{radosavovic2020designing} for the server-side model. We avoided overly large CNN architectures (like ResNet101 or Transformers) on the near-sensor side, thus simplifying real deployment. Note that our near-sensor backbones are significantly lighter (MobileNetV3-level complexity) compared to the server-side fusion model, keeping the computational overhead small. In late fusion models, we used multiple linear layers with the ReLU activation function.

\subsection{Dataset}
We evaluated our proposed method using the SynDrone dataset~\cite{rizzoli2023syndrone}, a public multi-modal dataset designed for urban classification applications for UAVs. We selected RGB images and depth maps to train our fusion-based intelligent sensing system. The dataset includes seven coarse classes, such as roads, construction, etc. and each coarse class contains several subclasses. Among these coarse classes, we decided to have the vehicle coarse class as our object of interest where we consider frames containing such objects of interest as a frame of interest. The vehicle coarse class contains 6 subclasses: Car, Truck, Bus, Train, Motorcycle, and Bicycle. In our evaluation, we assume a scenario of the edge-side or server-side fusion model conducting complex tasks, which is multi-label classification using the 6 subclasses. We took 80\% of the dataset as a train set and considered the other as a test set. Models are only trained with the train set and evaluated with the test set in all of our evaluations. Since our focus is to showcase how near-sensor multi-modal filtering can improve energy efficiency, we limit ourselves to this representative dataset, which already offers a variety of urban environment conditions and object classes.

\subsection{Evaluation of trade-off between quality loss and effective data generation}

\begin{figure}
    \centering    \includegraphics[width=0.9\linewidth]{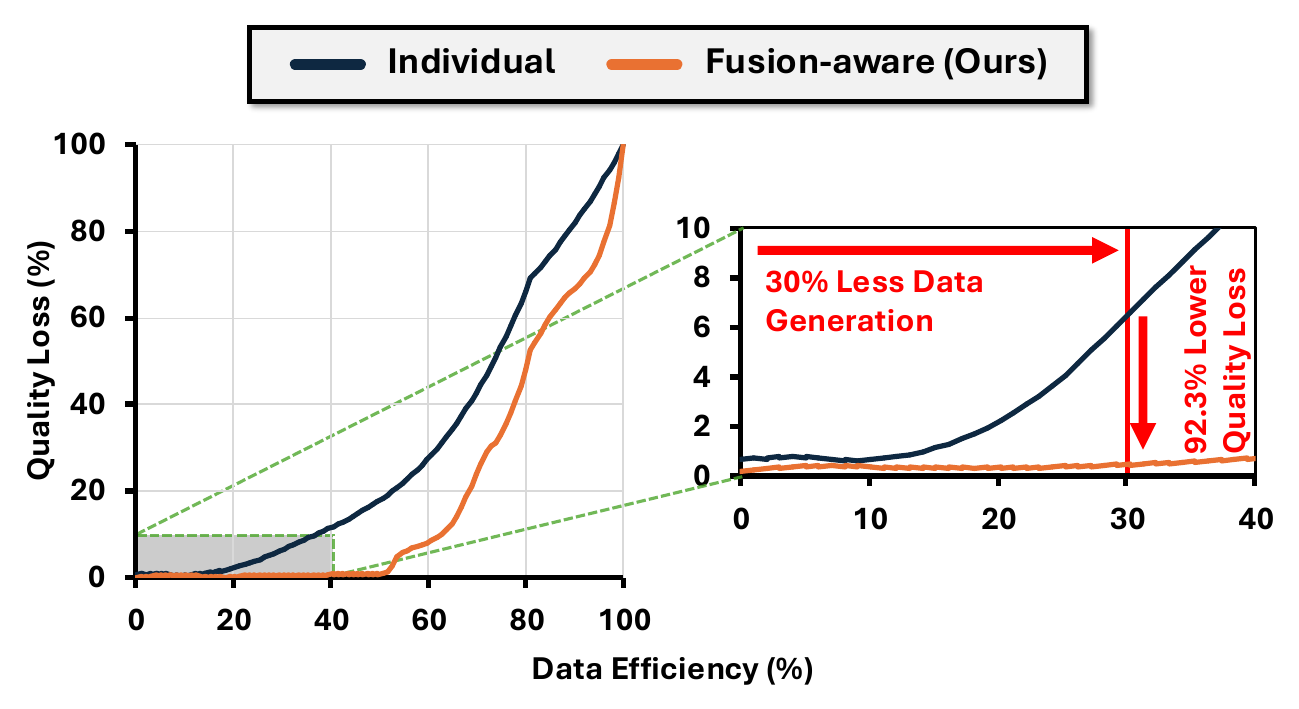}
    \vspace{-6mm}
    \caption{\small Trade-off relationship between data efficiency indicating the saved portion of the data in size and quality loss which is the performance drop rate of the server-side model when using filtered-out data.}
    \label{fig:tradeoff}
    \vspace{-8mm}
\end{figure}

To demonstrate the data selection efficiency of our near-sensor model compared to the previous intelligent sensing approach, we analyzed the trade-off between data filter-out rate and quality loss. This relationship is depicted in \autoref{fig:tradeoff}. We define the data filter-out rate, or data efficiency, as the proportion of data volume that is filtered out relative to the total data volume of all frames in the test set. Quality loss is assessed by the reduction in performance, specifically the macro F1 score, of the server-side fusion model when it processes the filtered data. From the theoretical perspective, each near-sensor model tries to approximate the server-side fusion output by leveraging the correlation discovered during training. The results reveal a distinct trade-off curve; our approach exhibits a steeper curve, achieving significantly higher data efficiency with substantially lower quality loss. Specifically, we observed a 92.3\% reduction in quality loss while targeting a 30\% reduction in data generation.

\subsection{Evaluation of edge fusion model compactization}

\begin{figure}
    \centering    \includegraphics[width=0.9\linewidth]{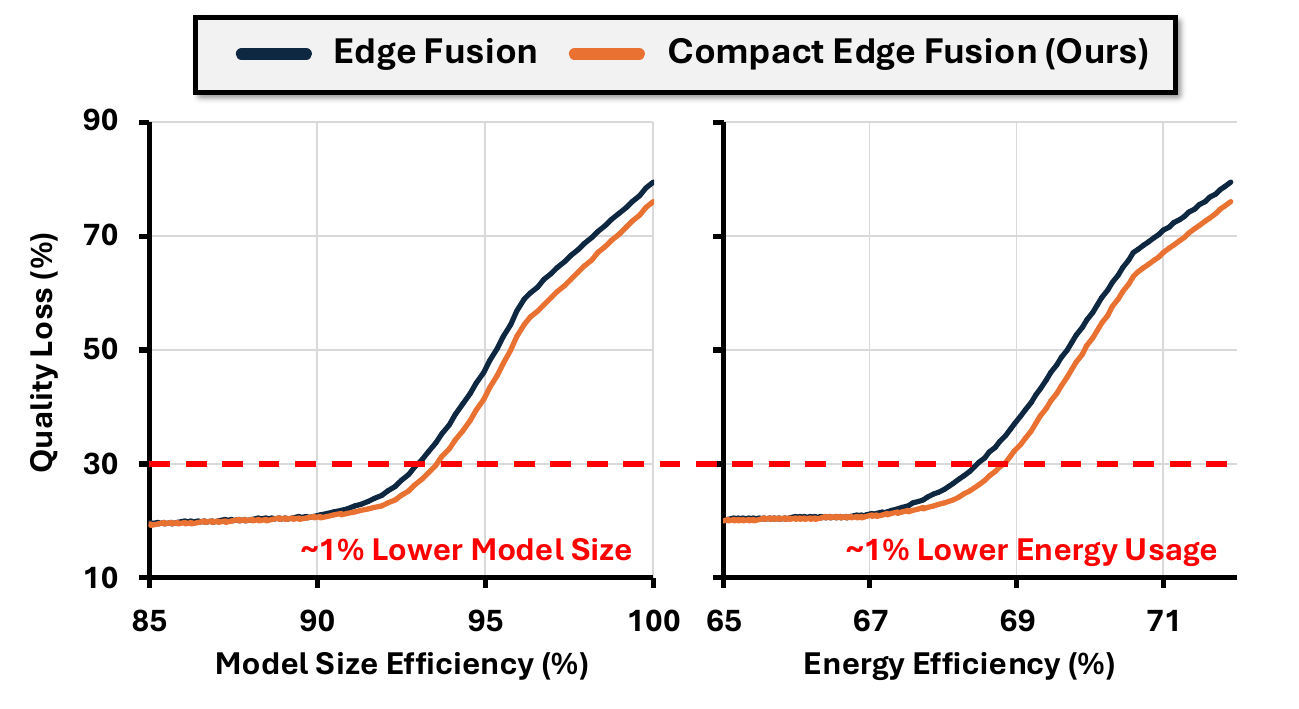}
    \vspace{-6mm}
    \caption{\small Quality loss by different edge-side fusion model sizes in terms of the number of trainable parameters compared to the server-side fusion model (left) and when converting the model size into energy efficiency compared to the energy consumption of server-side fusion model in Joule (right).}
    \label{fig:compactedgemodel}
    \vspace{-8mm}
\end{figure}

In this evaluation, we demonstrate the effectiveness of our proposed compact edge fusion model, designed to perform multi-label classification tasks similarly to the server-side fusion model. Unlike the latter, our compact model utilizes prediction scores from the near-sensor models during the late fusion stage. For comparison, we trained a baseline fusion model that does not incorporate near-sensor model scores yet maintains the same number of trainable parameters as our compact model.

\autoref{fig:compactedgemodel} presents the evaluation results, showcasing the quality loss relative to the server-side model across varying levels of model size efficiency, which indicates the reduction in trainable parameters compared to the server-side fusion model. Additionally, the figure includes a conversion of model size to energy efficiency. To estimate energy consumption, we considered a scenario using the Raspberry Pi 3 Model B as the edge computing device. Based on previous research~\cite{velasco2018performance}, we converted the model size to energy usage (in Joules) for different model sizes running on a Raspberry Pi.

While the parameter count of the edge model does not explode for two modalities, we note that in principle more modalities could increase total parameters. However, our approach partially mitigates this effect by shifting some complexity to near-sensor modules that individually remain lightweight and can be parallelized. The results reveal that our compact model achieves approximately 1\% lower model size and energy consumption compared to a typical edge fusion model, with a permissible quality loss of above 20\%. Although this represents a modest improvement in energy efficiency, it suggests that further optimizations, particularly in leveraging directional information from near-sensor models to an edge-side model, could enhance the compactness of the edge model without full integration.

\subsection{Energy efficiency on end-to-end system}

Inspired by a previous study on transmission energy consumption~\cite{nirjon2013auditeur}, we undertake a comprehensive evaluation to assess the end-to-end energy consumption of our proposed intelligent sensing framework. In this analysis, we focus on a scenario involving near-sensor models with data transmission to the cloud for processing by a server-side fusion model where the server is equipped with 13th Gen Intel(R) Core(TM) i9-13900KF with NVIDIA RTX 4070 GPU. We exclude considerations of edge-side modeling, focusing instead on real-time network communication to the server-side model. Also, we tested with three different scenarios where each has a different probability of observing FoI of 1\%, 5\%, and 10\%. To enhance the system's overall energy efficiency, developing an energy-optimized near-sensor model is crucial. We implemented our models on Google Edge TPU, utilizing ASIC acceleration, and estimated the TPU's energy consumption by measuring latency and referencing recent studies for average power consumption metrics~\cite{ni2022online}.

\autoref{fig:pie_breakdown} provides an energy consumption breakdown for four different approaches: the conventional method, which naively transmits all data; compressive sensing, specifically using Bit Depth Compression (BDC) to reduce energy costs, as highlighted in recent real-time applications~\cite{hwang2023lossless}; the previous filtering-out approaches~\cite{yun2024hypersense, huang2024plug}; and our proposed method employing fusion-aware near-sensor models developed through the three-step training process. The energy costs normalized to the conventional approach are central to each distribution. In case of the compression method, unlike the conventional approach, shows a significant reduction in communication costs. However, it still sends out all data, cannot exploit the low FoI probability scenarios, and fails to effectively reduce total energy consumption. On the other hand, our method demonstrates the most substantial energy savings, achieving up to 33$\times$ lower energy consumption than the conventional approach, and outperforming other state-of-the-art methods, which show at most a 22$\times$ reduction.

While the proposed system appears complex, it can be feasibly deployed with off-the-shelf edge TPUs or GPUs. The near-sensor models are light enough for real-time use, and adding new sensors only requires adding lightweight binary classifiers. Hence, we believe that the complexity is manageable, and the energy savings justify this design.

\section{Conclusions}\label{sec:conclusions}
We introduced \textsc{FusionSense}, a fusion-aware intelligent sensing framework that moves part of multimodal decision making to the sensing front-end. Using a three-step recipe—server-side fusion learning, \emph{filter-out-safe} supervision, and a compact edge fusion model fed by near-sensor scores—our system makes \emph{pre-fusion} keep/drop decisions that co-optimize compute and communication. On camera–LiDAR experiments, \textsc{FusionSense} achieves up to 33$\times$ lower energy at 1\% FoI and 11$\times$ at 10\%, with a 92.3\% reduction in quality loss at 30\% data reduction and roughly 1.5$\times$ higher energy savings than the best prior filter. The design scales \emph{linearly} with sensor count and runs on common accelerators without specialized hardware. Future work will extend to additional modalities and tasks and explore online adaptation under dynamic energy and deadline constraints.

\begin{acks}
This work was supported in part by the DARPA Young Faculty Award, the National Science Foundation (NSF) under Grants \#2127780, \#2319198, \#2321840, \#2312517, and \#2235472, the Semiconductor Research Corporation (SRC), the Office of Naval Research through the Young Investigator Program Award, and Grants \#N00014-21-1-2225 and N00014-24-1-2547, Army Research Office Grant \#W911NF2410360. Additionally, support was provided by the Air Force Office of Scientific Research under Award \#FA9550-22-1-0253.
\end{acks}


\bibliographystyle{ACM-Reference-Format}
\bibliography{mybibliography}










\end{document}